\documentclass[lettersize,journal]{IEEEtran}
\usepackage{amsmath,amsfonts}
\usepackage{algorithmic}
\usepackage{algorithm}
\usepackage{array}
\usepackage[caption=false,font=normalsize,labelfont=sf,textfont=sf]{subfig}
\usepackage{textcomp}
\usepackage{stfloats}
\usepackage{url}
\usepackage{verbatim}
\usepackage{graphicx}
\usepackage{cite}
\usepackage{orcidlink}
\begin{document}
	
\title{Depth as Points: Center Point-based Depth Estimation}
	
\author{
	Zhiheng Tu~\orcidlink{0009-0008-6353-6642}, Xinjian Huang~\orcidlink{0000-0001-8956-7326}, Yong He, Ruiyang Zhou, \\
	Bo Du~\orcidlink{0000-0002-0059-8458}~\IEEEmembership{Senior Member, IEEE}, Weitao Wu~\orcidlink{0000-0003-0400-6832}
\thanks{This work has been submitted to the IEEE for possible publication. Copyright may be transferred without notice, after which this version may no longer be accessible.}
\thanks{Zhiheng Tu, Yong He, Ruiyang Zhou and Weitao Wu are with the School of Mechanical Engineering, Nanjing University of Science and Technology, Nanjing 210094, China}
\thanks{Xinjian Huang is with the School of Cyber Science and Engineering, Nanjing University of Science and Technology, Nanjing 210094, China}
\thanks{Bo Du is with the School of Computer Science , Wuhan University, Wuhan, China.}
}
\maketitle
	
	\begin{abstract}
		The perception of vehicles and pedestrians in urban scenarios is crucial for autonomous driving. This process typically involves complicated data collection, imposes high computational and hardware demands. To address these limitations, we first develop a highly efficient method for generating virtual datasets, which enables the creation of task- and scenario-specific datasets in a short time. Leveraging this method, we construct the virtual depth estimation dataset VirDepth, a large-scale, multi-task autonomous driving dataset. Subsequently, we propose CenterDepth, a lightweight architecture for monocular depth estimation that ensures high operational efficiency and exhibits superior performance in depth estimation tasks with highly imbalanced height-scale distributions. CenterDepth integrates global semantic information through the innovative Center FC-CRFs algorithm, aggregates multi-scale features based on object key points, and enables detection-based depth estimation of targets. Experiments demonstrate that our proposed method achieves superior performance in terms of both computational speed and prediction accuracy. The code and datasets are available at \url{https://github.com/Elodie2001/CenterDepth}.
	\end{abstract}
	
	
	\section{Introduction}
	\IEEEPARstart{U}{rban} environment is essential for autonomous navigation and situational awareness\cite{ignatious2023analyzing,dong2023applications}.
	Accurately detecting and locating dynamic objects like pedestrians and vehicles in 3D space is critical for perceiving the surroundings and planning paths\cite{wevj16010020,ai5030061}. This involves object detection combined with depth estimation, which is vital in fields such as autonomous driving, intelligent transportation, and automated surveillance\cite{wang2020pseudolidarvisualdepthestimation,brazil2019m3drpnmonocular3dregion}.
	
	\begin{figure*}[h]
		\centering
		\includegraphics[width=\textwidth]{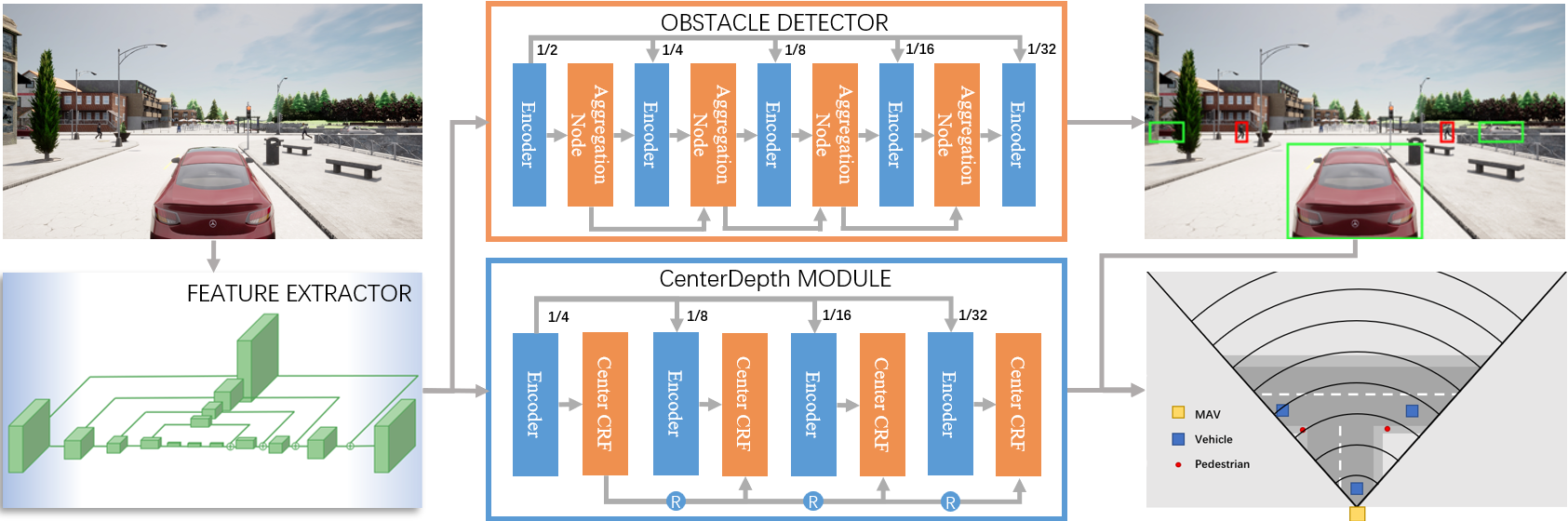}
		\caption{Overview of the Proposed System: The architecture consists of an obstacle detection network and a depth prediction module. The CenterDepth module performs localized depth prediction based on the object center predicted by the object detection task. The two tasks are jointly learned, with a shared feature extraction layer. Consequently, as the model receives target center information from the detector, it achieves improved accuracy in depth prediction.}
		\label{fig_1}
	\end{figure*}
	Deep learning methods have advanced significantly in this area, but they require large amounts of data for training\cite{liu2021groundawaremonocular3dobject,Schon_2021}. Creating datasets for specific tasks and scenarios is necessary, which traditionally involves equipping vehicles\cite{geiger2013vision,geyer2020a2d2} or drones\cite{yang2020reverse,lyu2020uavid} equipped with multiple sensors like LiDAR, stereo cameras, depth sensors, and RGB-D cameras\cite{6004839,6385934,doi:10.1177/0278364912455256} to collect data in urban settings. This process requires extensive post-processing for data consistency, including cleaning, error correction, and synchronization, making it time-consuming and expensive\cite{Caesar_2020_CVPR,Huang_2018_CVPR_Workshops,geyer2020a2d2}.
	
	To address these challenges, we propose a novel approach that leverages simulation for efficient dataset collection. The automated dataset collection method based on CARLA simulator\cite{pmlr-v78-dosovitskiy17a}. By invoking the 3D meshes of vehicles and pedestrians in a virtual environment, and through the projection of the intrinsic and extrinsic parameters calibrated by the sensors, the coordinates (x, y, z) of the eight vertices of the 3D bounding box and the center point of the cube are mapped into the camera coordinates to generate 2D annotations and depth data. Through the preset route settings in UE4, the ego car continuously runs in the simulated city to collect data. With the parameters including lighting conditions, weather, the number of vehicles and pedestrians in the city, and image resolution are settled, this function can generate a dataset containing more than 20,000 images in only 4 hours. This method allows us to autonomously collect data across various urban environments, simulating autonomous driving scenarios that form the foundation of our virtual dataset, virDepth.
	
	The dataset includes eight urban scenes of different sizes and styles, as well as various lighting and weather conditions. Each scene contains 200 vehicles of different types, 100 pedestrians and 5,000 RGB images, semantic images, and point cloud files. The annotations include both the 3D and 2D information of the objects.
	
	Monocular simultaneous localization and mapping (SLAM) techniques have been widely used for environmental perception\cite{pumarola2017pl,bruno2021lift,chung2023orbeez} through single-camera pose estimation. However, due to the lack of direct depth information, these methods must infer relative depth from motion, leading to scale ambiguity, making it hard to determine the absolute size of objects. Additionally, in environments with few textures or features, such as when small distant objects blend into the background, it's challenging to find enough feature points for accurate localization and mapping.
	
\begin{figure*}[!t]
		\centering
		\includegraphics[width=\textwidth]{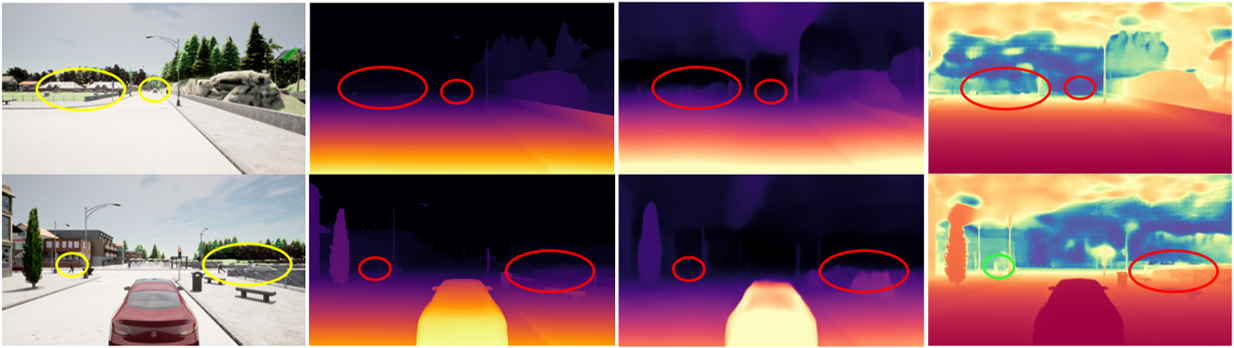}
		\caption{The depth prediction results obtained using traditional depth estimation methods are shown. From left to right, each column represents the original images, the results from DepthAnything, Monodepth and DepthAnythingV2. Red indicates that the depth of the target is no longer different from the background depth, and green indicates that the depth of the target area is successfully predicted. As observed from the images, the target within the yellow circle in the original image becomes indistinguishable from the background in the traditional depth estimation methods due to its small size, leading to prediction failure.}
		\label{fig_2}
	\end{figure*}

	Current depth estimation methodologies face inherent trade-offs between accuracy, efficiency, and practicality in autonomous driving scenarios. Multi-sensor fusion techniques achieve high precision through cross-modal supervision\cite{wan2022multi,ali2019multi}, yet they demand large-scale, meticulously synchronized datasets and specialized hardware configurations\cite{zhang2024occfusion}. These requirements escalate costs and limit scalability in dynamic urban environments. On the other hand, monocular depth estimation methods\cite{godard2019digging,yang2024depthv1} prioritize computational efficiency but suffer from some critical limitations. Generating pixel-wise depth maps for entire images imposes excessive computational overhead, conflicting with the real-time constraints of autonomous vehicles\cite{liu2021groundawaremonocular3dobject}. As shown in Fig.\ref{fig_2}, for distant small objects, blurred boundaries and sparse texture exacerbate scale ambiguity, leading to detection failures. While recent efforts integrate SLAM with depth estimation\cite{bruno2021lift}, scale drift persists in textureless regions, and hybrid frameworks often lack task-specific optimization for driving scenarios\cite{chung2023orbeez}.
	
	To tackle these intertwined challenges, we propose CenterDepth, a unified framework that synergizes targeted object detection with localized depth regression. Unlike conventional monocular methods that compute global depth map, CenterDepth focuses on object-centric depth estimation.
	
	Unlike conventional global depth estimation techniques, CenterDepth estimates depth by focusing on the center point of the object. Our approach incorporates a target detection network based on the CenterNet\cite{zhou2019objects} framework, which predicts the object's center to serve as the anchor point for depth estimation within the detection box. This strategy eliminates the need for generating a global depth map, significantly reducing computational demands.
	
	Furthermore, we propose the Center FC-CRFs mechanism to address scale ambiguity and challenges associated with textureless surfaces. By facilitating information propagation within the regions corresponding to vehicles and pedestrians using a local Fully-connected Conditional Random Fields (FC-CRFs) approach, Center FC-CRFs reduces the computational demands associated with  FC-CRFs while achieving high-precision monocular depth estimation across a large-scale range. Experiment show that our method present 1.308\% absolute error for target in range 0-50 meters, while 2.070\% for target in range 150-200 meters.
	
	The entire system first detects the target by predicting its center point. Then, local depth estimation is performed based on the target's central point, and the spatial position of the target is determined through projection transformation. An overview of the system is presented in Fig.\ref{fig_1}.

	To evaluate our system, we trained and validated the model using the virDepth and tested its adaptability with various backbones. Experimental results demonstrate that CenterDepth achieves excellent depth prediction accuracy while operating efficiently on the onboard computing systems of autonomous vehicles, making it a promising solution for real-time applications in urban environments.

	The contributions are summarized as follows:
	\begin{itemize}
		\item{To address the challenges of high costs and difficulties in data collection for autonomous driving scenarios, we propose a CARLA-based data collector and construct a fine-grained dataset tailored for vehicle and pedestrian detection and depth prediction in urban autonomous driving environments.}
		\item{To overcome the inefficiencies of existing depth estimation methods and poor performance on long-distance targets, we introduce a center-point regression-based depth prediction module, CenterDepth, and the innovative CenterCRFs mechanism, enabling the propagation of local features within a large-scale range.}
		\item{Experiments validate the effectiveness of our method through comparisons with traditional depth prediction techniques, and we further verify the generalizability and robustness of CenterDepth by integrating it into various backbone architectures.}
	\end{itemize}
	
	This paper is organized as follows. Section 2 reviews related work on depth estimation in autonomous driving scenarios. Section 3 presents the network architecture and algorithm implementation of our proposed method. Section 4 details the process of constructing the virDepth using the aforementioned data collector. Section 5 outlines the experimental procedures and results.
	
	\section{Related Work}
	We reviewed depth estimation methods in autonomous driving scenarios based on different sensor types: monocular-LiDAR fusion, stereo camera-LiDAR fusion, and monocular depth prediction.
	
	Monocular-LiDAR fusion\cite{ma2018sparse,tang2020learning,chen2019learning,park2020non,li2019dense} is a method based on depth completion algorithms. This approach propagates the high-precision but sparse points from the LiDAR sensor and aims to densely complete the depth map by leveraging image information. Recent deep learning-based methods have significantly improved the quality of depth estimation. These methods fuse information by projecting the point cloud into the image domain and feeding the sparse depth map as input into the network. On the other hand, \cite{chen2019learning} introduced an intermediate fusion strategy in feature space, where features are extracted from each modality and fused in the image feature space using projected point features. Although this approach achieves a certain level of prediction accuracy, depth completion often fails in regions not covered by the point cloud or where the points are too sparse.
	
	Stereo camera-LiDAR fusion\cite{park2018high,park2019high,maddern2016real,zhu2022vpfnet,wang20193d} has recently been shown to further improve the accuracy of depth estimation by utilizing additional sensory information. \cite{park2018high} refined the disparity map from stereo cameras using sparse depth maps. Although this method improves the quality of depth estimation, the fusion is performed in the 2D image domain, which is insufficient to maintain the metric accuracy of the point cloud for long-distance depth estimation. \cite{wang3d} introduced the idea of input fusion and volumetric normalization based on sparse disparity. However, this method struggles to accurately estimate depth beyond more than 150 meters.
	
	In response, we propose a monocular camera-based depth prediction method, significantly reducing the cost of equipment, data collection, and operation compared to LiDAR and stereo devices, enabling autonomous driving on lightweight platforms.
	
	With the continuous development of deep learning, monocular camera-based 3D object detection has emerged as another environment perception approach. This method can directly predict the 3D information of vehicles within the field of view\cite{wang2021fcos3d,liu2020smoke}, but its primary task is to regress the 3D bounding boxes and orientations of objects, often neglecting depth prediction. For targets beyond 80 meters, depth prediction almost entirely fails. Additionally, these methods generate large amounts of prediction data, have complex models, and exhibit low computational efficiency, making them unsuitable for real-time detection in autonomous driving. Similarly, traditional global depth estimation methods\cite{yang2024depthv1,yang2024depthv2,godard2017unsupervised}, which lack direct depth information from monocular images, are prone to scale ambiguity. In environments with few textures or sparse features, these models struggle to accurately estimate the depth of distant targets.
	
	To address these challenges, we adopt a center-point-based depth prediction approach, inferring the depth of the target from the predicted object center. This approach does not require precise object boundary identification and performs well in scenarios where features are blurred or missing.
	
	Before the advent of deep learning, monocular depth estimation was a challenging task. It was difficult to infer pixel depth solely from the pixels themselves, as global context of the entire image needed to be considered. To address this, discriminative training of Markov Random Fields (MRF) was employed to establish the relationship between multi-scale local and global features. Graphical models, such as MRFs\cite{saxena2008make3d} and Conditional Random Fields (CRFs)\cite{wang2015depth}, have been shown to be effective in depth estimation. CRFs enhance depth estimation and edge preservation by utilizing both local and global contextual information from the image. For instance, \cite{xu2017multi,ricci2018monocular} applied CRFs to depth map optimization, effectively eliminating local inconsistencies through the integration of global information. However, the high computational demand of CRFs in full-field optimization often results in real-time issues. Additionally, the global smoothing strategy of CRFs can reduce the accuracy of local depth, especially in scenes with complex boundaries.
	
	In this work, inspired by \cite{yuan2022neural}, we propose a Center FC-CRFs module based on the neural window  FC-CRFs, using global features as input. By treating the target key point as a feature anchor and the region within the object detection bounding box as a local window, we use the Center FC-CRFs module to aggregate global features and optimize target depth prediction.
	
	\section{Network Architecture}
	We regress the object center point using a heatmap, taking this key point as a global feature anchor to aggregate contextual semantic information and regress the target depth, as shown in Fig.\ref{fig_3}. The detailed network architecture and algorithmic implementation are described in the following sections.
	\subsection{Center Point Regression}
	Let the input image to be $I\in R^{W\times H\times3}$ , denoted by $W$(width) and $H$(height). We use several different fully convolutional encoder-decoder architectures (such as ResNet or Hourglass networks) to extract multi-scale features and regress keypoint heatmaps to locate the centers $\begin{pmatrix}x_{c},y_{c}\end{pmatrix}$ of vehicles and pedestrians as well as their sizes $h$, $w$. Let $N$ denote the size of the extracted feature map $F\in\mathbb{R}^{N\times N\times C}$, and $C$ represent the number of channels in the feature space. The feature map retains the spatial and semantic information necessary for detecting vehicles and pedestrians across various scales and estimating their depth.
	We define the vehicle and pedestrian regions as follows:
	\begin{figure*}[!t]
	\centering
	\includegraphics[width=\linewidth]{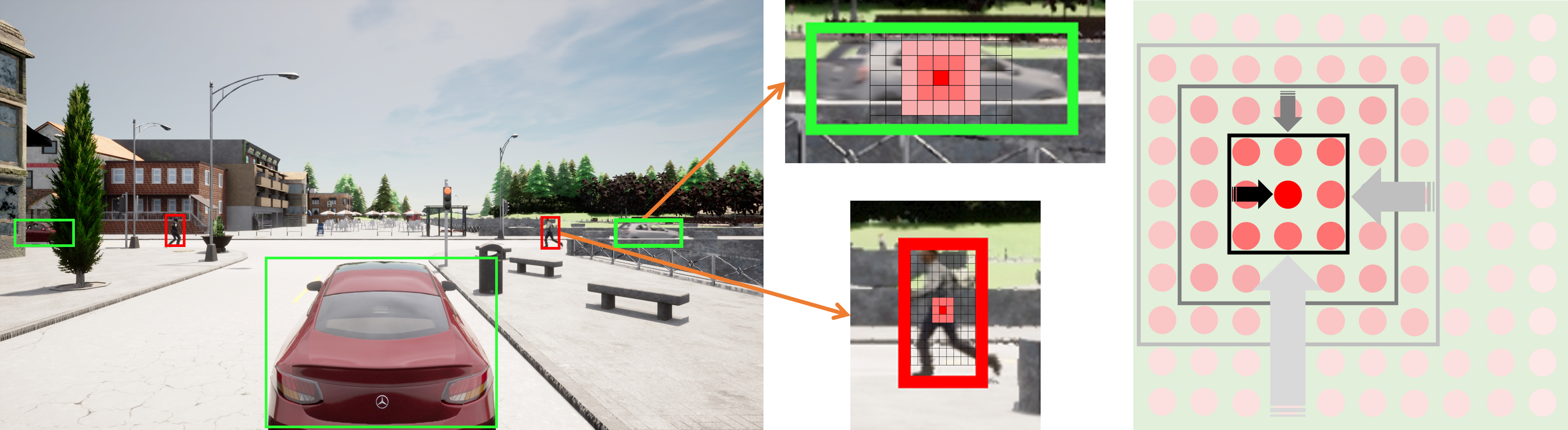}	\caption{The left panel illustrates the keypoint prediction of objects using heatmaps, where the intensity of each pixel represents the confidence of the corresponding keypoint location. The right panel depicts the principle of CenterCRFs, which adaptively allocates weights to pixels based on their distance to the keypoint. This mechanism enables keypoint-based feature aggregation by assigning higher weights to pixels in the vicinity of the keypoint, thereby enhancing the semantic consistency and spatial coherence of features around the target's central location.}
	\label{fig_3}
\end{figure*}	
	\begin{equation}
		\label{eq1}
		\mathbb{R}=\left\{\begin{pmatrix}x,y\end{pmatrix}|\begin{vmatrix}x-x_c\end{vmatrix}\le\frac{w}{2},\begin{vmatrix}y-y_c\end{vmatrix}\le\frac{h}{2}\right\}
	\end{equation}
	
	\subsection{Center FC-CRFs}
	The depth $D\bigl(x_{i},y_{i}\bigr)$ of each pixel $(x_{i},y_{i}\bigr)$ within the region $\mathbb{R}$ is modeled as the fully connected conditional probability distribution of the vehicle and pedestrian depth values, with the center point of the vehicle and pedestrian serving as the anchor point for all depth values. This point represents the geometric center of the entire vehicle or pedestrian. We optimize this using the FC-CRFs model, connecting all nodes within the region to the center point, thereby propagating positional feature information to accurately estimate the depth distribution of the vehicle and pedestrian center points.It is defined as follows:
	
	\begin{equation}
		\label{eq2}
		\begin{pmatrix}x_c,y_c\end{pmatrix}=\arg\max_{(x,y)\in R}H\begin{pmatrix}x,y\end{pmatrix}
	\end{equation}
	
	\noindent where $B\bigl(x,y\bigr)$ represents the heatmap extracted from the feature map, and denotes the confidence distribution of the center point.
	
	In the FC-CRFs model, we treat the center point as a strongly constrained point, establishing connections with other pixels in the vehicle and pedestrian regions through pairwise potentials $\psi_{p}$. The depth of these pixels in the field of view is defined to vary with the depth of the center point as follows:
	
	\begin{align}
		\label{eq3}
		\psi_p&\Big(D\big(x_i,y_i\big),D\big(x_c,y_c\big)\Big) \nonumber \\ &=\omega_{i,c}\cdot\Big\|D\big(x_i,y_i\big)-D\big(x_c,y_c\big)\Big\|^2
	\end{align}
	
	\noindent where $\omega_{i,c}$ represents the weight between each point $(x_{i},y_{i}\bigr)$ and the center point, which we define as the weight based on feature similarity:
	
	\begin{equation}
		\label{eq4}
		\omega_{i,c}^{\text{feat}}=\exp\left(-\frac{\left\|\text{f}\left(x_i,y_i\right)-\text{f}\left(x_c,y_c\right)\right\|^2}{2\sigma_f^2}\right)
	\end{equation}
	
	\noindent where $\mathrm{f}(x_{c},y_{c})$ and $\mathrm{f}(x_{i},y_{i})$ are the feature vectors of the center point and the remaining pixels, respectively, and $\sigma_{f}$  is the standard deviation parameter controlling the influence of feature similarity. When the difference between the feature vectors is smaller, the weight is larger, and vice versa. The energy function $E(D)$ of our Center FC-CRFs model is defined as follows:
	
	\begin{align}
		\label{eq5}
		E\left(D\right)=&\sum\psi_{u}\left(D\left(x,y\right)\right)  \nonumber\\ &+\sum\psi_{p}\left(D\left(x_{i},y_{i}\right),D\left(x_{\epsilon},y_{\epsilon}\right)\right)
	\end{align}
	
	\noindent where $\psi_{u}$ represents the unary potential of the depth likelihood at each point, and $\psi_{p}$ denotes the pairwise potential for spatial coherence between each point and the center point within the region. The center point $(x_{c},y_{c}\bigr)$ acts as the central node, connected to the boundary points of the region, ensuring smooth depth estimation within the vehicle and pedestrian boundaries while maintaining sharp edge transitions. The final regional $D_{_R}$ depth map is obtained by minimizing this energy function.

	\section{Virtual Depth Estimation Datasets}
	\begin{figure*}[!t]
		\centering
		\subfloat[]{\includegraphics[width=2.3in]{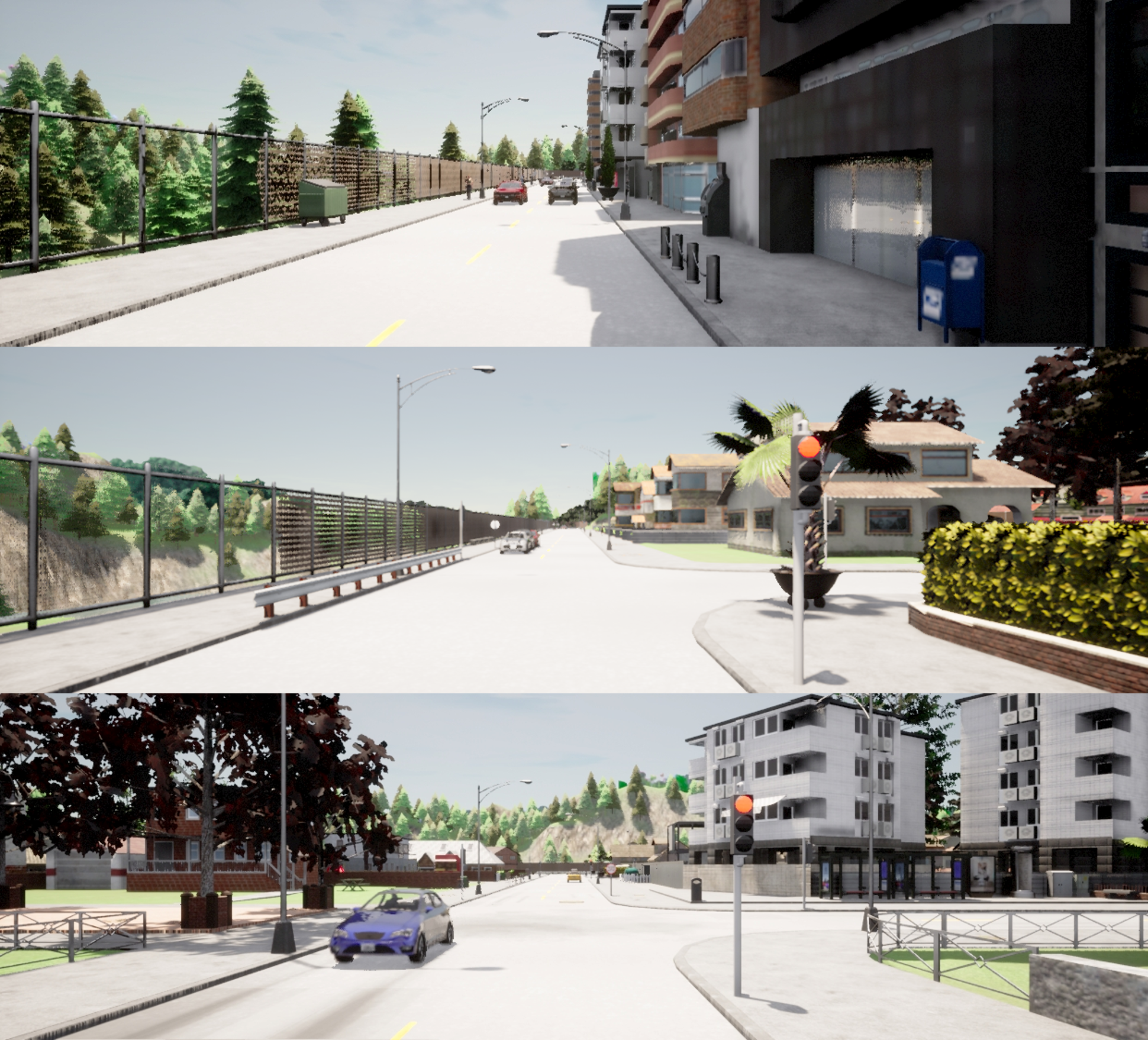}
			\label{fig_4_1}}
		\hfil
		\subfloat[]{\includegraphics[width=2.3in]{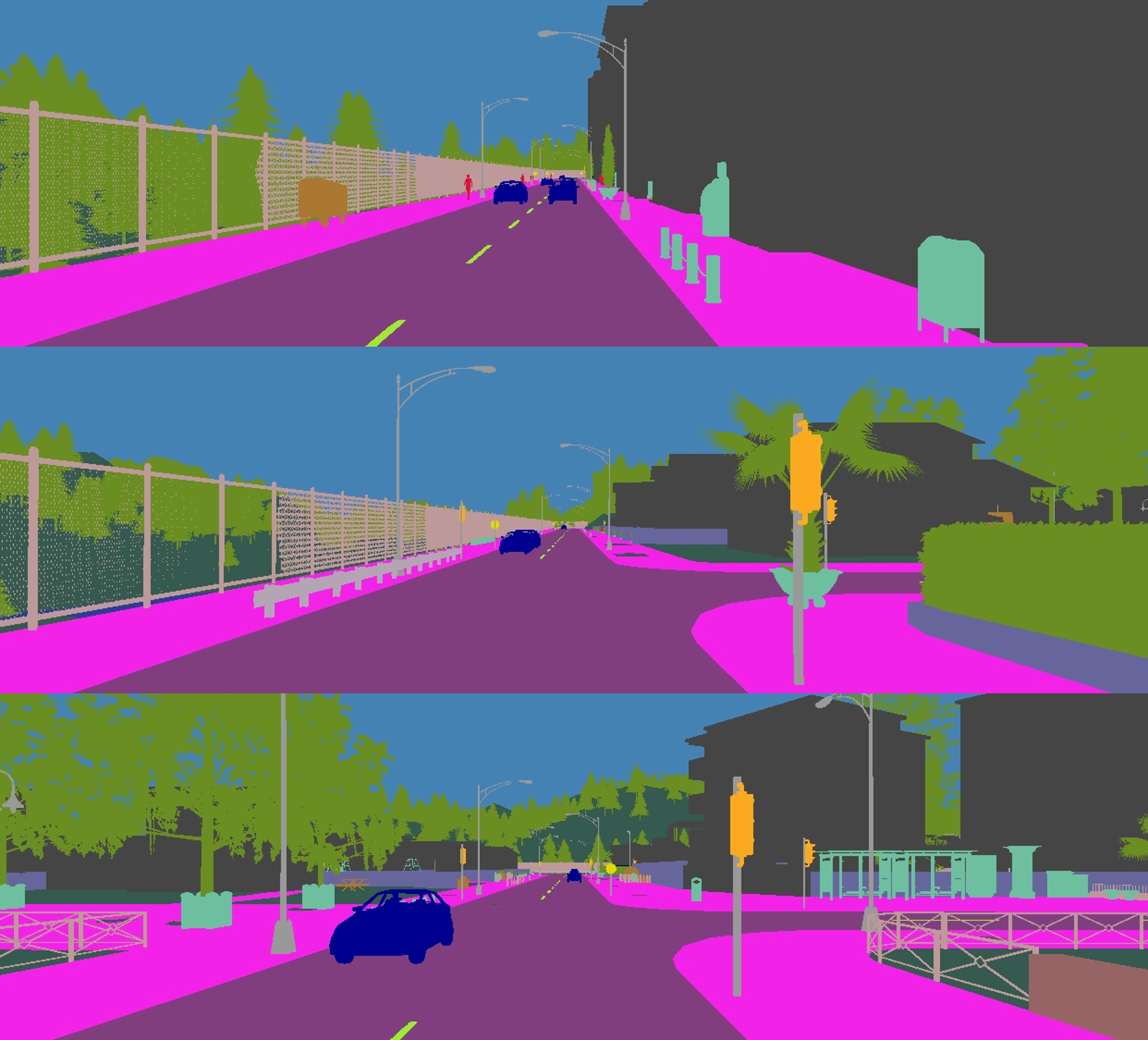}
			\label{fig_4_2}}
		\hfil
		\subfloat[]{\includegraphics[width=2.3in]{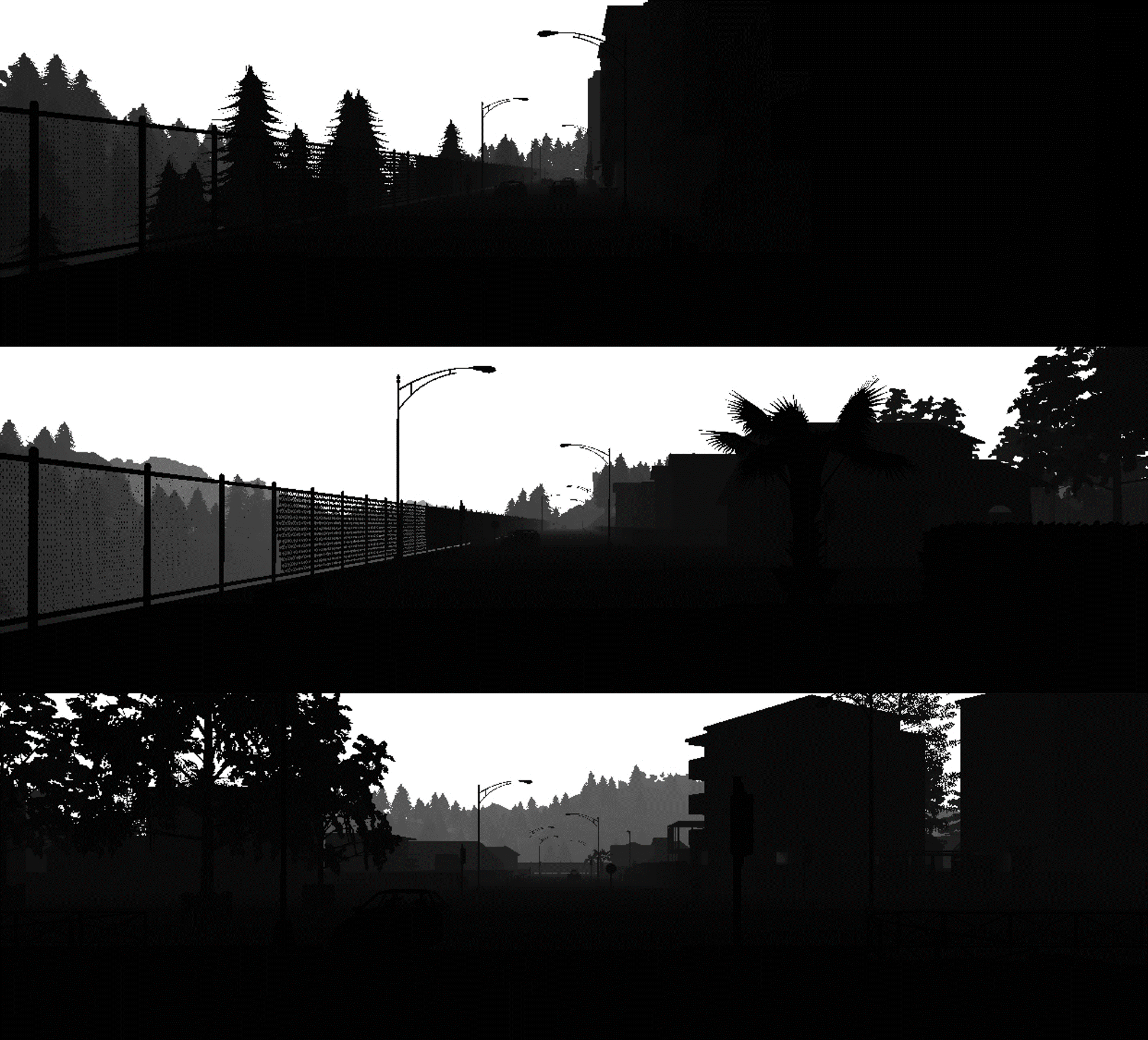}
			\label{fig_4_3}}
		\caption{A schematic of the virDepth. (a) shows the RGB images, (b) shows the semantic segmentation images, and (c) shows the depth images.}
		\label{fig_4}
	\end{figure*}
	\begin{figure}[!t]
		\centering
		\includegraphics[width=\linewidth]{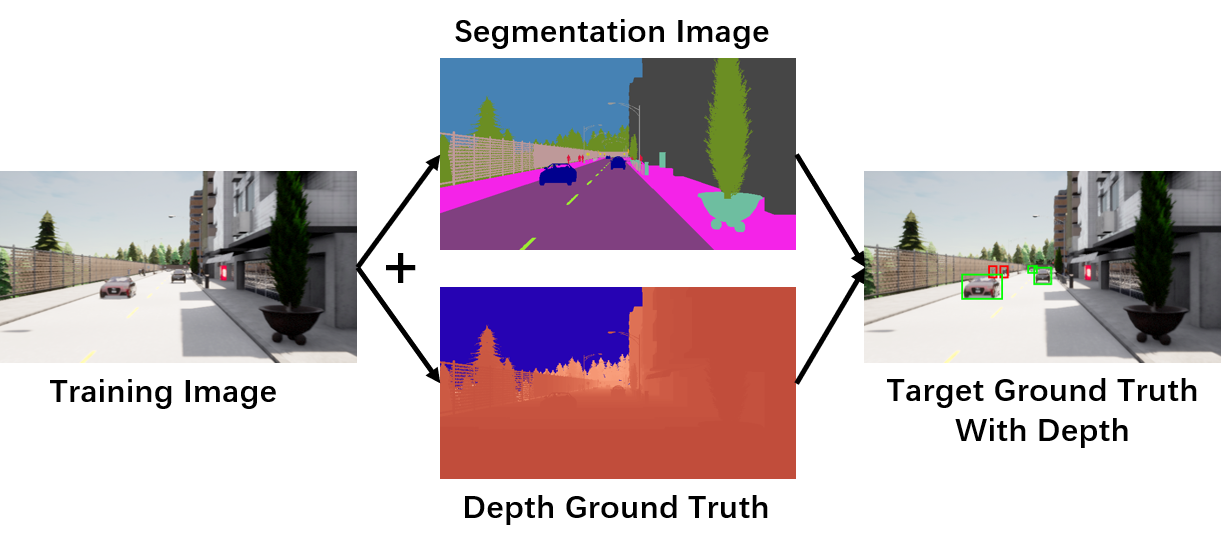}
		\caption{The training images were selected based on the given semantic segmentation maps, identifying the regions corresponding to vehicles and pedestrians. The depth information for these regions was extracted from the depth ground truth maps. We evaluate the vehicle and pedestrian targets within a 100-meter range.}
		\label{fig_5}
	\end{figure}
	We established a data collection framework using Carla and Unreal Engine 4, generating a virtual datasets comprising more than 50,000 images across five highly realistic urban scenes of varying sizes. This datasets is tailored for detecting various vehicles and pedestrians in urban environments. 
	
	We implemented an autonomous driving program for the ego vehicle, enabling it to navigate autonomously in the created urban scenarios. During operation, the ego vehicle simultaneously collects data from three types of cameras installed directly in front of it: an RGB optical camera, a depth camera, and a semantic segmentation camera. These three cameras share the same spatial position, orientation, focal length, and data collection frequency. All their internal and external parameters are set to be the same as those in the KITTI\cite{geiger2013vision}. Therefore, the datasets is automatically synchronized and aligned. A schematic of virDepth is shown in Fig.\ref{fig_4}.
	
	Our method extract the eight vertices $\mathbf{P}_{\omega}=\begin{pmatrix}X_{\omega},Y_{\omega},Z_{\omega}\end{pmatrix}$ of the 3D bounding boxes of vehicles and pedestrians, and transform these world coordinates into the camera coordinate system using the extrinsic matrix:
	\begin{equation}
		\label{eq6}
		\mathrm{P}_{c}=\mathrm{W}2\mathrm{C}\cdot\mathrm{P}_{\omega}
	\end{equation}
	
	\noindent where $\mathrm{W2C=}[R_{3\times3},t_{3\times1}]$, $R$ is a $3\times3$ rotation matrix, $t$ is a $t_{3\times1}$ translation vector. Next, the 3D points in the camera coordinate system are projected onto the image plane using the intrinsic matrix $K$ (also referred to as the projection matrix):
	
	\begin{equation}
		\label{eq7}
		p_i=\mathrm{K}\cdot\mathrm{P}_c
	\end{equation}
	
	\noindent where the intrinsic matrix $K$ typically takes the following form:
	
	\begin{equation}
		\label{eq8}
		\text{K=}\begin{bmatrix}f_x&0&c_x\\0&f_y&c_y\\0&0&1\end{bmatrix}
	\end{equation}
	
	where $f_x$ and $f_y$ are the focal lengths of the camera multiplied by the size of each pixel, and $c_x$, $c_y$ are the coordinates of the principal point (the projection of the optical axis on the image).Subsequently, we project them onto the image plane coordinate system via the intrinsic matrix. By normalizing the coordinates to eliminate depth information, we obtain the 2D bounding box image coordinates. 
	
	Subsequently, we simulated the ego car's operation in each urban environment, automatically collecting data from the sensors and generating 2D and 3D annotations based on the physical model structures and world coordinates of vehicles, pedestrians, and bicycles within the field of view. Additionally, using data from the semantic segmentation and depth cameras, we performed post-processing on the segmentation labels to classify objects into cars, vans, trucks, bicycles, and pedestrians. An overview of the data generation pipeline is illustrated in Fig.\ref{fig_5}. Targets exceeding 200 meters in distance or with a high occlusion ratio were filtered out, resulting in the generation of ground truth labels for the detection and depth of vehicles, bicycles, and pedestrians within a 200-meter range.
	
	In this function, we have configured multiple customizable parameters, including lighting conditions, weather, scenario types, categories, quantities, and distributions of targets, sensor types, resolutions, and data collection rates. When the data collection program runs, the ego car continuously navigates through the virtual city, while vehicles and pedestrians in the environment remain in constant motion. Using this approach, researchers can collect over 20,000 distinct data samples within 4 hours. Compared with the public datasets Virtual KITTI 2\cite{cabon2020virtual}, our method enables flexible construction of more tailored datasets for different tasks and scenarios.
	
	\section{Experiments}
	\begin{figure*}[!t]
		\centering
		\includegraphics[width=\textwidth]{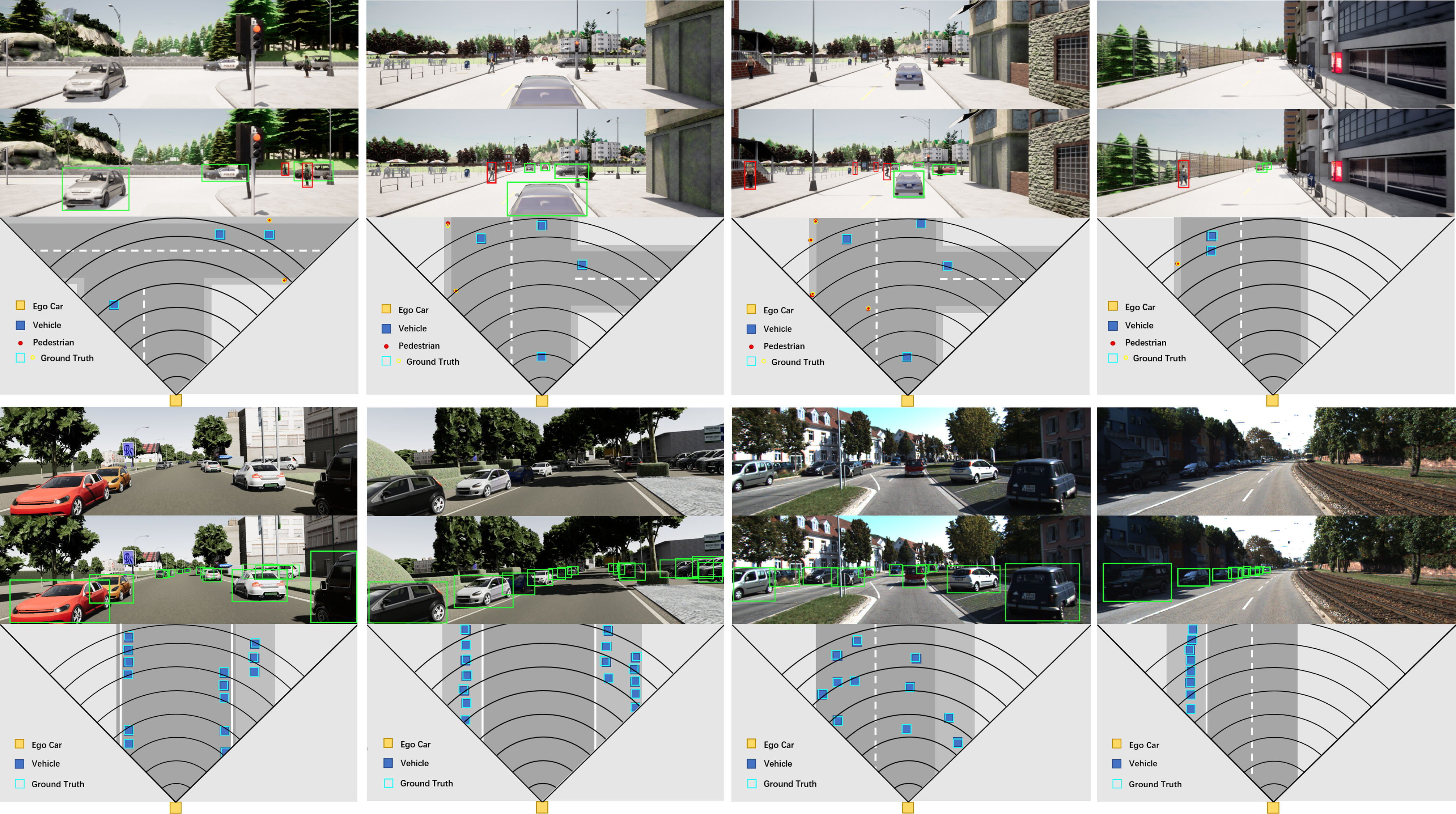}
		\caption{The selected results. The first row shows the results on the virDepth, while the first two columns in last row show the results on the Virtual KITTI datasets, and the last two columns show the results on KITTI dataset. In every figure, first row represents the original images, the second row shows the detection results, and the third row presents the BEV (Bird's Eye View) results obtained by projecting the depth predicted based on the detection results.}
		\label{fig_6}
	\end{figure*}
	
	\begin{table*}[!t]
		\centering
		\caption{Comparing with the state-of-art on the virDepth, Virtual KITTI 2 and KITTI-Depth test set.\label{tab:table1}}
		\resizebox{\textwidth}{!}{
			\renewcommand{\arraystretch}{1.5}
			\begin{tabular}{l|c|c c c c c |c c c c c |c c c c c}
				\hline
				\multicolumn{2}{c|}{Datasets} & \multicolumn{5}{c|}{\textit{virDepth}} & \multicolumn{5}{c|}{\textit{Virtual KITTI 2}} &\multicolumn{5}{c}{\textit{KITTI-Depth}}\\
				\hline
				Methods & Backbone & $\delta_{1}$ & $\delta_{2}$ & $\delta_{3}$ & $MRE$ & $RMSE$ & $\delta_{1}$ & $\delta_{2}$ & $\delta_{3}$ & $MRE$ & $RMSE$ & $\delta_{1}$ & $\delta_{2}$ & $\delta_{3}$ & $MRE$ & $RMSE$\\
				\hline
				DepthAnythingV2\cite{yang2024depthv2} &-& 0.832 & 0.946 & 0.968 & 0.104 & 13.155 & \textbf{0.914} & 0.957 & 0.993 & 0.095 & 6.456 & 0.925 & 0.964 & 0.983 & 0.113 & 8.524\\
				MonoDepth2\cite{godard2017unsupervised} &-& 0.811 & 0.937 & 0.979 & 0.130 & 16.802 & 0.878 & 0.962 & 0.974 & 0.106 & 9.923 & 0.867 & 0.937 & 0.983 & 0.125 & 10.020\\
				CenterDepth* &HGL104& 0.804 & 0.897 & 0.982 & 0.114 & 6.802 & 0.822 & 0.841 & 0.986 & 0.105 & 6.923 & 0.735 & 0.860 & 0.881 & 0.133 & 8.654\\
				CenterDepth &HGL104& \textbf{0.989} & \textbf{1.000} & \textbf{1.000} & \textbf{0.041} & \textbf{2.106} & 0.901 & \textbf{0.986} & \textbf{1.000} & \textbf{0.065} & \textbf{4.261} & \textbf{0.963} & \textbf{0.972} & \textbf{1.000} & \textbf{0.076} & \textbf{5.357}\\
				\hline
			\end{tabular}		
		}
	\end{table*}
	
	\begin{table*}[!t]
		\centering
		\caption{Comparing with the state-of-art on the virDepth, Virtual KITTI 2, KITTI-3D test set.\label{tab:table2}}
		\resizebox{\textwidth}{!}{
			\renewcommand{\arraystretch}{1.5}
			\begin{tabular}{l|c|c c c c c |c c c c c |c c c c c}
				\hline
				\multicolumn{2}{c|}{Datasets} & \multicolumn{5}{c|}{\textit{virDepth}} & \multicolumn{5}{c|}{\textit{Virtual KITTI 2}} &\multicolumn{5}{c}{\textit{KITTI-3D}}\\
				\hline
				Methods & Backbone & $\delta_{1}$ & $\delta_{2}$ & $\delta_{3}$ & $MRE$ & $RMSE$ & $\delta_{1}$ & $\delta_{2}$ & $\delta_{3}$ & $MRE$ & $RMSE$ & $\delta_{1}$ & $\delta_{2}$ & $\delta_{3}$ & $MRE$ & $RMSE$\\
				\hline
				SMOKE\cite{liu2020smoke} &R101& 0.821 & 0.945 & 0.951 & 0.089 & 9.324 & 0.854 & 0.890 & 0.973 & 0.087 & 8.401 & 0.831 & 0.943 & 0.951 & 0.101 & 9.992\\
				FCOS3D\cite{wang2021fcos3d} &R101& 0.750 & 0.864 & 0.986 & 0.067 & 5.837 & 0.845 & \textbf{0.934} & 0.964 & \textbf{0.063} & 4.412 & 0.840 & \textbf{0.947} & 0.983 & 0.070 & 7.310\\
				CenterDepth* &R101& 0.715 & 0.778 & 0.947 & 0.074 & 7.453 & 0.737 & 0.832 & 0.985 & 0.066 & 7.258 & 0.667 & 0.738 & 0.967 & 0.095 & 8.020\\
				CenterDepth &R101& \textbf{0.983} & \textbf{0.996} & \textbf{1.000} & \textbf{0.053} & \textbf{2.688} & \textbf{0.863} & 0.932 & \textbf{0.992} & 0.068 & \textbf{2.437} & \textbf{0.848} & 0.920 & \textbf{1.000} & \textbf{0.044} & \textbf{2.804}\\
				\hline
				DD3D\cite{park2021pseudo} &V2-99& 0.759 & 0.864 & \textbf{1.000} & 0.058 & 5.837 & \textbf{0.845} & 0.892 & 0.980 & 0.063 & 6.412 & 0.825 & 0.947 & \textbf{0.993} & 0.082 & 7.310\\
				BEVFormer\cite{li2203bevformer} &V2-99& \textbf{0.874} & 0.936 & 0.993 & \textbf{0.054} & 5.103 & 0.801 & \textbf{0.916} & \textbf{0.992} & \textbf{0.057} & 6.201 & \textbf{0.853} & \textbf{0.966} & 0.980 & \textbf{0.063} & 6.624\\
				CenterDepth* &V2-99& 0.715 & 0.838 & 0.980 & 0.083 & 7.453 & 0.778 & 0.832 & 0.985 & 0.069 & 7.258 & 0.631 & 0.753 & 0.849 & 0.106 & 8.020\\
				CenterDepth &V2-99& 0.862 & \textbf{0.956} & 0.997 & 0.076 & \textbf{4.554} & 0.830 & 0.875 & 0.934 & 0.060 & \textbf{5.892} & 0.819 & 0.902 & 0.967 & 0.074 & \textbf{6.382}\\
				\hline
			\end{tabular}		
		}
	\end{table*}
	\begin{table*}[!t]
		\centering
		\caption{Comparison of Model Complexity and Real-Time Performance Across Different Backbone Networks.\label{tab:table3}}
		\resizebox{\linewidth}{!}{
			\renewcommand{\arraystretch}{1.5}
			\begin{tabular}{l|c|c c|c c c c c|c c c c c}
				\hline
				\multicolumn{4}{c|}{Environment} & \multicolumn{5}{c|}{On single RTX 4090 GPU} & \multicolumn{5}{c}{On Jetson Orin NX 16GB}\\
				\hline
				Backbone &Resolution& FLOPs(B) & Params(M)& Time(ms) & $\delta_{1}$ & MRE & MAE & RMSE & Time(ms) & $\delta_{1}$ & MRE & MAE & RMSE\\
				\hline
				Hourglass-104 &512$\times$512& 42.28 & 123.72& 15.6 & \textbf{0.989} & \textbf{1.13\%} & \textbf{1.394} & \textbf{2.106} & 83.2 & \textbf{0.841} & \textbf{6.59\%} & \textbf{5.976} & \textbf{6.615}\\
				DLA-34 &512$\times$512& 14.84 & 29.41& 6.1 & 0.962 & 2.77\% & 2.648 & 5.205 & 32.7 & 0.774 & 7.35\% & 6.305 & 7.812 \\
				ResNet-18 &512$\times$512& 2.17 & 6.43& \textbf{3.4} & 0.909 & 3.01\% & 3.415 & 6.436& \textbf{25.0} & 0.725 & 11.24\% & 9.353 & 13.058 \\
				ResNet-101 &512$\times$512& 31.57 & 69.02& 9.5 & 0.983& 2.31\% & 2.834 & 2.688 & 49.8 & 0.792& 8.93\% & 7.623 & 6.960\\
				VoVNet-99 &512$\times$512& 35.09 & 85.33 & 11.4 & 0.862& 7.64\% & 8.065 & 4.873& 44.7 & 0.740& 9.07\% & 9.489 &  11.512\\
				\hline
			\end{tabular}
		}
	\end{table*}
	\begin{table}[!t]
		\centering
		\caption{Results of subdivided regions on virDepth. R1 to R4 indicate subdivided regions 0-50meters, 50-100meters, 100-150meters, 150-200meters.\label{tab:table4}}
		\resizebox{\columnwidth}{!}{
			\renewcommand{\arraystretch}{1.5}
			\begin{tabular}{l|c| c c c c}
				\hline
				Method & Backbone & $MAE_{R1}$ & $MAE_{R2}$ & $MAE_{R3}$ & $MAE_{R4}$ \\
				\hline
				DepthAnythingv2 &-& 0.625 & 1.875 & 21.750 & 36.269 \\
				Monodepth2 &-& 0.734 & 2.326 & 28.496 & 41.677 \\
				CenterDepth* &H104& 0.612 & 2.849 & 6.478 & 13.537 \\
				CenterDepth &H104& \textbf{0.491} & \textbf{1.036} & \textbf{1.990} & \textbf{3.351} \\
				\hline
				SMOKE &R101& 0.825 & 1.248 & 6.795 & 12.203 \\
				FCOS3D &R101& 0.798 & 1.540 & 4.374 & 10.679 \\
				CenterDepth* &R101& 0.541 & 1.953 & 3.440 & 8.148 \\
				CenterDepth &R101& \textbf{0.327} & \textbf{1.149} & \textbf{2.279} & \textbf{3.624} \\
				\hline
				DD3D &V2-99& \textbf{0.304} & 1.347 & 3.325 & 11.483 \\
				BEVFormer &V2-99& 0.316 & \textbf{1.132} & 2.573 & 6.354 \\
				CenterDepth* &V2-99& 0.541 & 1.953 & 3.440 & 8.148 \\
				CenterDepth &V2-99& 0.359 & 1.149 & \textbf{2.982} & \textbf{4.407} \\
				\hline
			\end{tabular}	
		}
	\end{table}
	
	\subsection{Datasets}
	
	We conduct experiments on Self-constructed datasets virDepth and three challenging public autonomous driving datasets, namely Virtual KITTI 2 ,KITTI-Depth, KITTI-3D.
	
	\textbf{virDepth} contains 5 scenes with roughly 15,000 images. To match the dataset with KITTI, we use the proposed dataset generation function to create a new dataset of 1242×375 resolution with the same camera parameters and poses. It includes RGB, depth, and segmentation images with complete labels of 2D and 3D bounding boxes to fit different models’ tasks. Notably, virDepth contains over 180,000 targets evenly distributed within the range of 0–200 meters. On average, each image contains three targets within every 50-meter interval. This dataset provides large-scale content for autonomous driving scenarios.
	
	\textbf{Virtual KITTI 2, KITTI-Depth, KITTI-3D} supports multiple vision tasks, including stereo vision, optical flow, visual odometry, 3D object detection, tracking, road and lane detection, and semantic segmentation. We obtain the ground truth of the target center depth by extracting the three - dimensional position of the target and calculating its distance from the ego car. The $\delta_{1}$, $\delta_{2}$, $\delta_{3}$, Mean Relative Error (MRE), Mean Absolute Error (MAE), and Root Mean Square Error (RMSE) are used to evaluate the model's prediction performance for the target center depth. Set the $\delta$ threshold as 1.10.
	
	For the training of CenterDepth on 3D detection datasets, we obtain 2D bounding boxes either by projecting 3D bounding boxes onto the image plane or by directly using the 2D bounding boxes included in the datasets. We then extract object positions from the 3D information and compute depth information as labels.
	
	For full-image depth estimation method, we propose two methods—Center and Seg—to extract target depth from the results for comparative experiments.
	
	In the Center approach, the detector in our method predicts the centers of vehicles and pedestrians in the RGB images, infers the image coordinates of these center points, and then extracts the corresponding depth information from the depth maps generated by the other methods. 
	
	In the Seg approach, we use the semantic segmentation images from the dataset as masks, extracting the depth information from the vehicle and pedestrian regions in the depth maps, and taking the average to obtain the depth prediction for the target.
	
	\subsection{Implementation Details}
	We adopt three types of backbone: Hourglass104 that initialized from CenterNet\cite{zhou2019objects}, ResNet101-DCN\cite{dai2017deformableconvolutionalnetworks} that initialized from FCOS3D\cite{wang2021fcos3d} and VoVnet-99\cite{lee2019energygpucomputationefficientbackbone} that initialized from DD3D\cite{park2021pseudo}. We train CenterDepth on the virDepth in five different urban scenes. All methods are trained on an NVIDIA RTX 4090 GPU. Training is conducted using the Adam optimizer, with StepLR as the learning rate scheduler. The initial learning rate is set to 0.000125, which remains constant for the first 90 epochs, and then decreases tenfold at the 90th and 180th epochs until convergence.
	
	To verify the model's performance for targets at different distances, we divide the entire target area into four parts: 0 - 50 meters, 50 - 100 meters, 100 - 150 meters, and 150 - 200 meters to highlight the performance comparison.

	\subsection{Depth estimation results}
	\begin{figure*}[!t]
		\centering
		\includegraphics[width=\textwidth]{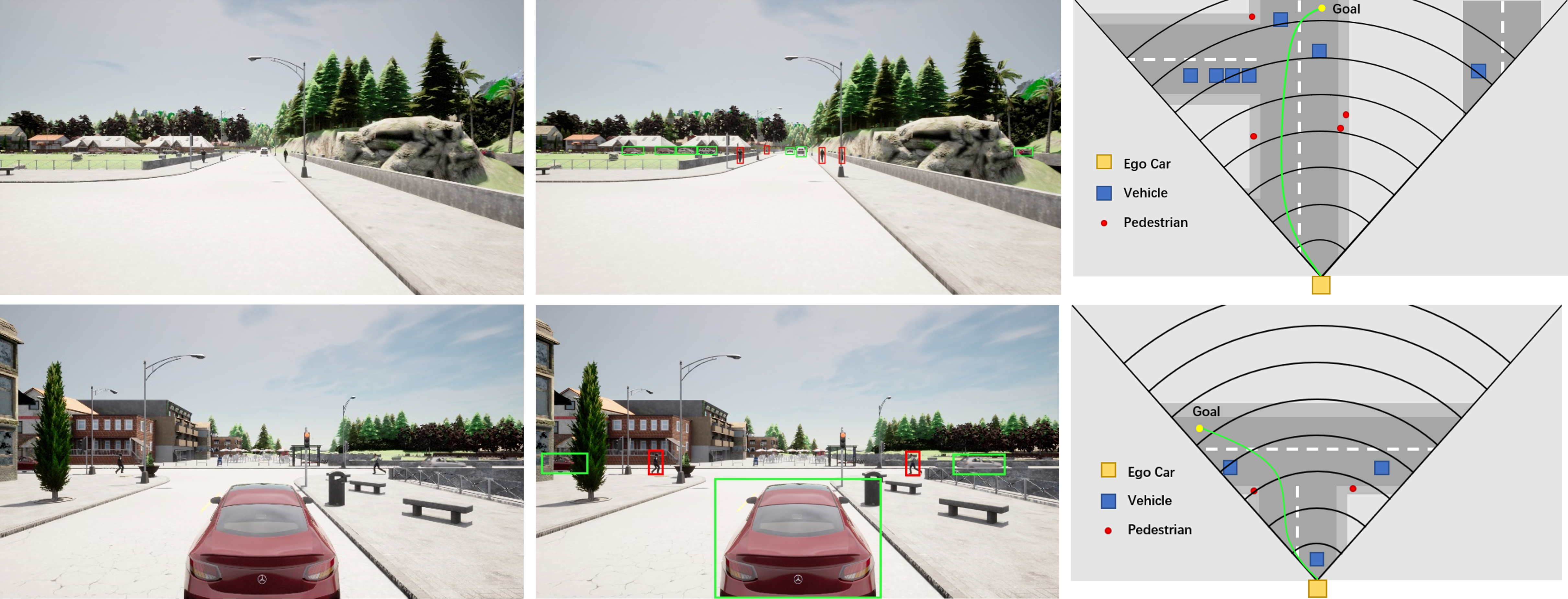}
		\caption{Path planning results based on CenterDepth. As shown in the figure, the first column represents the original images, the second column displays the detection results for vehicles and pedestrians, and the last column presents the path planning results by the ego car based on the aforementioned detection and depth information. From the images, it can be observed that our system combines both sources of information to accurately locate obstacles from a BEV perspective and perform path planning accordingly.}
		\label{fig_7}
	\end{figure*}
	\begin{figure*}[!t]
		\centering
		\includegraphics[width=\textwidth]{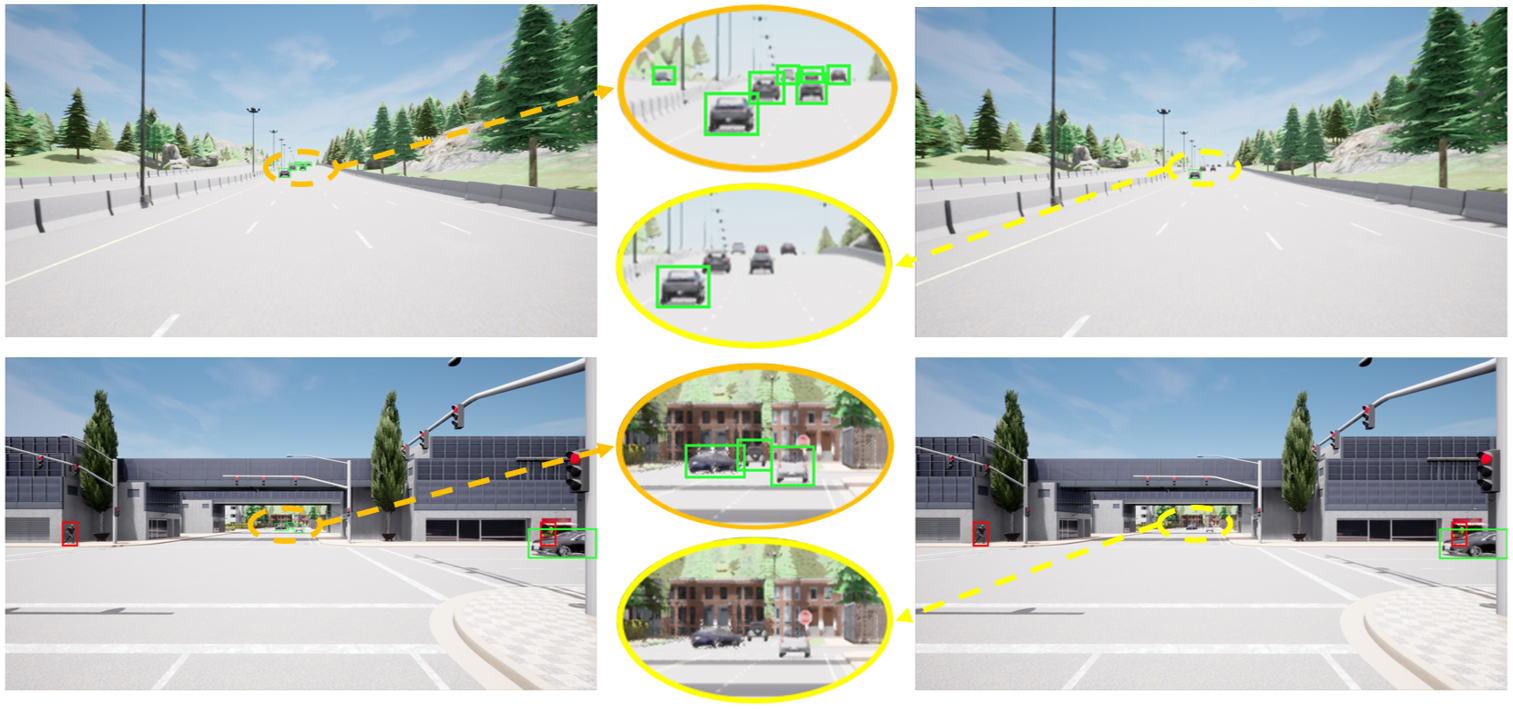}
		\caption{The figure shows the results of CenterDepth (left) and YOLOv8 (right). Our method demonstrates superior performance in detecting small targets at distances of 100-200 meters.}
		\label{fig_8}
	\end{figure*}
	In the comprehensive evaluation across multiple datasets and distance intervals, the proposed CenterDepth model demonstrates superior target center depth prediction performance and unique large-scale balancing capabilities. Selected results are shown in Fig.\ref{fig_6}. As shown in the core metrics of monocular depth prediction show in table.\ref{tab:table1} and 3D detection show in table.\ref{tab:table2}, CenterDepth significantly outperforms comparative methods on the virDepth, Virtual KITTI 2, and KITTI-Depth/KITTI-3D test sets. Taking the virDepth as an example, CenterDepth with the HGL104 backbone achieves a $\delta_{1}$ of 0.989 in monocular depth prediction tasks (an 18.9\% improvement over the next-best method, DepthAnythingV2), with an MRE as low as 0.041 and an RMSE of only 2.106—representing reductions of 60.6\% and 84.0\% compared to the second-place methods, respectively. This highlights its high-precision localization capability for target center depths. In 3D detection scenarios, CenterDepth with the R101 backbone achieves a $\delta_{1}$ of 0.983 and an RMSE of 2.688 on the virDepth, reducing errors by 71.2\% and 54.0\% compared to SMOKE and FCOS3D, respectively. This validates its depth prediction advantages through the fusion of semantic segmentation and center extraction techniques.
	
	The model’s large-scale balancing capability is particularly evident in the distance-subdivided analysis show in table.\ref{tab:table4}. When the target area is divided into four sub-intervals—0–50 meters (R1), 50–100 meters (R2), 100–150 meters (R3), and 150–200 meters (R4)—CenterDepth achieves the lowest 
	MAE values across all intervals under different backbone configurations. For example, with the HGL104 backbone, its MAE values in R1–R4 are 0.491, 1.036, 1.990, and 3.351, respectively, representing reductions of 21.4\%–90.8\% compared to the corresponding values of the depth prediction method DepthAnythingV2 (0.625, 1.875, 21.750, 36.269) and significant superiority over 3D detection methods such as BEVFormer (R4 MAE= 6.354). Notably, in the long-distance interval of 150–200 meters (R4), CenterDepth’s MAE
	is 60.5\%–57.0\% lower than that of the same-backbone CenterDepth* (baseline model), indicating that its targeted design effectively mitigates the accuracy degradation in long-distance depth prediction. This avoids the drastic performance decline with increasing distance observed in traditional methods (e.g., Monodepth2’s R4 MAE of 41.677).
	
	Cross-dataset comparisons further highlight CenterDepth’s generalization ability: it consistently achieves the lowest MREand RMSE on the Virtual KITTI 2 and KITTI-Depth test sets. In the complex scenarios of KITTI-3D, CenterDepth with the V2-99 backbone achieves an MAE of 4.407 in the R4 interval, representing improvements of 61.6\% and 30.7\% compared to DD3D (11.483) and BEVFormer (6.354), respectively. These results demonstrate that CenterDepth, through optimized geometric constraints on target center depth and multi-scale feature fusion, not only achieves high-precision prediction across all distance ranges but also exhibits significantly superior balancing performance in large-scale scenarios compared to existing methods. This provides a reliable technical foundation for autonomous driving and other tasks requiring long-range perception.
	
	In summary, our method outperforms current state-of-the-art models in accuracy, reliability, and computational efficiency. By achieving the lowest error rates and maintaining consistently high performance across both synthetic and real-world datasets, our approach establishes a new benchmark for monocular depth prediction. The notable improvements over previous methods highlight the practical effectiveness of our technique, particularly for applications in autonomous driving and robotics.
	
	\subsection{Efficiency Analysis}
	To evaluate the trade-off between inference efficiency and depth estimation accuracy, we compared the performance of CenterDepth with different backbone networks on single RTX 4090 gpu and the Nvidia Jetson NX 16GB, as summarized in table.\ref{tab:table3}. The results demonstrate a clear distinction in how each backbone balances speed and precision.
	
	In terms of inference efficiency, the ResNet-18 backbone achieved the fastest mean runtime (mRt) of 0.025 seconds, showcasing its lightweight architecture. However, this speed advantage came at the cost of significantly lower accuracy: it recorded the lowest $\delta_{1}$ (0.909), highest MRE (3.01\%), MAE (3.415), and second-highest RMSE (6.436) among all backbones. At the opposite end, the Hourglass-104 backbone set the accuracy benchmark with the highest $\delta_{1}$ (0.989), lowest MRE (1.13\%), MAE (1.394), and RMSE (2.106), but required a longer mRt of 0.038 seconds, 52\% slower than ResNet-18.
	
	Notably, the ResNet-101 backbone emerged as the optimal balance between efficiency and accuracy. With an mRt of 0.029 seconds, it was only 16\% slower than ResNet-18 yet significantly outperformed it in all accuracy metrics: $\delta_{1}$ was 8.1\% higher (0.983 vs. 0.909), MRE and MAE were 23.2\% and 16.9\% lower, respectively, and RMSE was reduced by 58.2\%. Compared to the accuracy-leading Hourglass-104, ResNet-101 achieved 99.4\% of its $\delta_{1}$ and 80.3\% of its RMSE while being 23.7\% faster. In contrast, DLA-34 and VoVNet-99 exhibited suboptimal performance: DLA-34 had moderate speed (0.032 s) but poor accuracy, and VoVNet-99 suffered from both the slowest mRt (0.042 s) and the worst accuracy (e.g., MRE=7.64\%, MAE=8.065), highlighting its inefficiency for real-world applications.
	
	These results demonstrate that CenterDepth effectively leverages backbone architectures to balance performance and efficiency. While Hourglass-104 prioritizes peak accuracy and ResNet-18 emphasizes extreme speed, ResNet-101 provides a practical compromise, achieving near-optimal accuracy with minimal runtime penalty. This flexibility allows CenterDepth to adapt to diverse computational environments without sacrificing critical depth estimation quality, underscoring its robustness as a versatile framework for efficient and accurate depth perception.
	
	\subsection{Limitations and Systematic Analysis}
	To validate effectiveness of CenterDepth, we integrate this module into the path planning system as an image processing and obstacle localization component. Using Unreal Engine 4 (UE4), we constructed a traffic intersection in an urban scene, where targets are mapped from image coordinates to world coordinates based on depth estimation results and the dimensions of detection bounding boxes. Cubes are placed at corresponding positions to represent human entities, and detected vehicles and pedestrians are visualized in a bird's-eye view (BEV), as shown in the figure.
	
	We use the urban scenes built for the datasets as path planning scenarios, setting the image capture location as the starting point and placing the endpoint randomly behind obstacles. The ego car maximum speed is set to 2m/s, and the images captured by the front RGB camera are used as input. From the captured inputs, we detect vehicles and pedestrians and estimate their depth. By reconstructing the scene, the obtained information is converted into BEV localization information for vehicles and pedestrians, as shown in the figure. We use A* to generate the path, as illustrated in the fig.\ref{fig_7}.
	
	Table.\ref{tab:table5} compares the detection performance of CenterDepth and YOLOv8s in the 150–200 meters region of the virDepth dataset, focusing on small/remote object detection accuracy, model complexity, and inference speed. CenterDepth achieves a 45.4\% mAP$_0.5:0.95$, significantly outperforming YOLOv8s by 15.4 percentage points. This large margin highlights its superiority in long-range small-object localization. The advantage stems from CenterDepth’s scale-adaptive feature aggregation and depth-aware context modeling, which enhance the representation of low-contrast, tiny targets. Selected results are shown in fig.\ref{fig_8}.
	
	\begin{table}[!t]
		\centering
		\caption{Detecion results of 150-200meters region on virDepth.\label{tab:table5}}
		\resizebox{\columnwidth}{!}{
			\renewcommand{\arraystretch}{1.5}
			\begin{tabular}{l|c| c c c c}
				\hline
				Method & size & mAP$_{0.5:0.95}$ & Params(M) & FLOPs(B) & Time(ms) \\
				\hline
				CenterDepth & 512$\times$512 & \textbf{45.4} & 14.8 & 29.4 & 6.1 \\
				YOLOv8s &640$\times$640& 30.0 & 11.2 & 28.6 & \textbf{4.3} \\
				\hline
			\end{tabular}	
		}
	\end{table}
	
	\section{Conclusion and Future Work} \label{sec:Conclusion and Future Work}
	In this work, we proposed the CenterDepth module, a novel end-to-end depth prediction framework with broad applicability, designed for depth estimation following object detection. We demonstrated its effectiveness in detecting various vehicles and pedestrians and estimating their depth on a self-constructed urban scene dataset. We also evaluated its applicability across different backbone networks. Furthermore, we validated the module’s ability to assist ego car in path planning using the provided information. In future work, we plan to further improve its operational efficiency to meet the demands of high-speed intelligent agent and enhance its robustness when dealing with large-scale variations of targets, which remains a key challenge for the real-world deployment of such algorithms.

	{
		\bibliographystyle{IEEEtran}
		\bibliography{references}
	}

	\vfill
	
\end{document}